\documentclass[letterpaper, 10 pt, conference]{ieeeconf}
\IEEEoverridecommandlockouts

\usepackage{xspace}
\usepackage{xcolor}
\usepackage{amsfonts}
\usepackage{graphics} %
\graphicspath{{figures/}}  %
\usepackage{epsfig} %
\usepackage{mathptmx} %
\usepackage{times} %
\usepackage{amsmath} %
\usepackage{amssymb}  %
\usepackage{ulem}
\usepackage{siunitx}
\usepackage{verbatim}
\usepackage{booktabs}
\usepackage{multirow}
\usepackage[space]{cite}
\usepackage[table]{xcolor} %
\definecolor{lightgreen}{RGB}{210,255,210}
\usepackage[colorlinks=true, linkcolor=blue, urlcolor=blue, citecolor=blue]{hyperref}

\usepackage[font=footnotesize]{caption}

\makeatletter
\let\origthebibliography\thebibliography
\def\thebibliography#1{%
  \origthebibliography{#1}%
  \scriptsize %
}
\makeatother

\newcommand{\method}{CRAFT\xspace}

\newcommand{\gt}{\text{g}}
\newcommand{\source}{\text{s}}
\newcommand{\generated}{\text{d}}
\newcommand{\lef}{\text{l}}
\newcommand{\rig}{\text{r}}

\definecolor{figgray}{rgb}{0.2, 0.2, 0.2}
\definecolor{figgreen}{rgb}{0.180, 0.494, 0.196}
\definecolor{figblue}{rgb}{0.082, 0.396, 0.753}
\definecolor{figlightred}{rgb}{1.0, 0.149, 0.0}
\definecolor{figdarkblue}{rgb}{0.188, 0.392, 0.729}

\title{\LARGE \bf
\method: Video Diffusion for Bimanual Robot Data Generation \\ 
}

\author{
  Jason Chen, I-Chun Arthur Liu, 
  Gaurav S. Sukhatme, Daniel Seita \\
  University of Southern California
}

\begin{document}
\twocolumn[{%
\renewcommand\twocolumn[1][]{#1}%
\maketitle
\vspace{-0.5em}
\begin{center}
    \centering
    \captionsetup{type=figure}
    \includegraphics[width=1.0\textwidth]{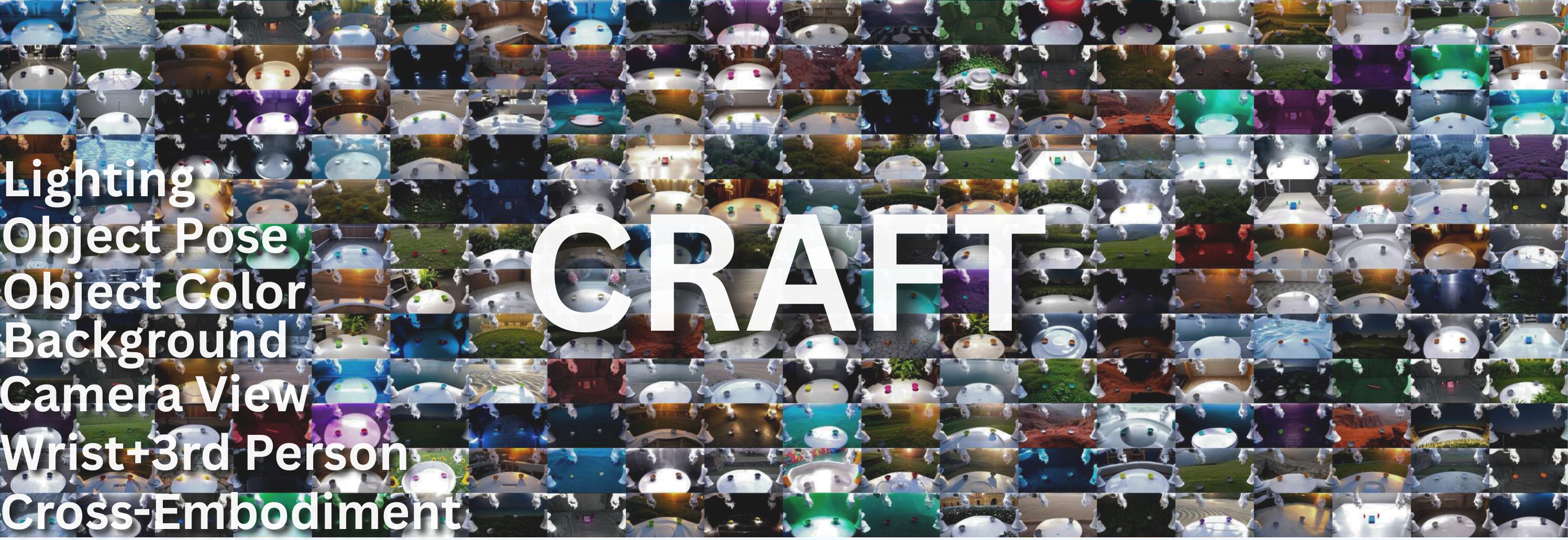}
    \vspace{-10pt}
    \caption{\textbf{CRAFT generates visually diverse bimanual manipulation demonstrations.} A mosaic of synthetic training videos produced by our video diffusion framework, spanning seven axes of visual variation: lighting conditions, object pose, object color, background environment, camera viewpoint, wrist and third-person perspectives, and cross-embodiment transfer. Starting from a small set of real-world teleoperation demonstrations, CRAFT synthesizes large-scale, photorealistic and temporally coherent demonstration datasets paired with action labels for coordinated dual-arm manipulation.}
    \label{fig:teaser}
\end{center}%
}]
\thispagestyle{empty}
\pagestyle{empty}

\begin{abstract}
Bimanual robot learning from demonstrations is fundamentally limited by the cost and narrow visual diversity of real-world data, which constrains policy robustness across viewpoints, object configurations, and embodiments. We present \underline{C}anny-guided \underline{R}obot D\underline{a}ta Generation using Video Di\underline{f}fusion \underline{T}ransformers (\method), a video diffusion-based framework for scalable bimanual demonstration generation that synthesizes temporally coherent manipulation videos while producing action labels. By conditioning video diffusion on edge-based structural cues extracted from simulator-generated trajectories, CRAFT produces physically plausible trajectory variations and supports a unified augmentation pipeline spanning object pose changes, camera viewpoints, lighting and background variations, cross-embodiment transfer, and multi-view synthesis. We leverage a pre-trained video diffusion model to convert simulated videos, along with action labels from the simulation trajectories, into action-consistent demonstrations. Starting from only a few real-world demonstrations, \method generates a large, visually diverse set of photorealistic training data, bypassing the need to replay demonstrations on the real robot (Sim2Real). Across simulated and real-world bimanual tasks, CRAFT improves success rates over existing augmentation strategies and straightforward data scaling, demonstrating that diffusion-based video generation can substantially expand demonstration diversity and improve generalization for dual-arm manipulation tasks.
Our project website is available at: \href{https://craftaug.github.io/}{https://craftaug.github.io/}.
\end{abstract}

\section{Introduction}

\begingroup
  \setlength{\skip\footins}{1pt}    %
  \setlength{\footnotesep}{2pt}     %
  \let\thefootnote\relax
  \footnotetext{%
    All authors are with the Thomas Lord Department of Computer Science at the University of Southern California, USA. Correspondence: \texttt{jchen567@usc.edu}.%
  }
\endgroup

Imitation learning coupled with large teleoperated datasets has enabled increasingly capable bimanual manipulation systems~\cite{black2024pi0visionlanguageactionflowmodel,intelligence2025pi05,liu2024rdt}. However, scaling these models to diverse embodiments, viewpoints, and task variations remains data-intensive, motivating scalable methods to expand demonstration diversity without additional real-world data~\cite{goldbergoldfashioned}.
While data augmentation has emerged as a promising strategy~\cite{tian2024vista,wang2026roboaug,yuan2025roboengineplugandplayrobotdata}, existing works focus on a subset of augmentations, such as augmenting only third-person~\cite{tian2024vista} or wrist-camera views~\cite{liu2025DCODA,zhang2024diffusionmeetsdagger}, or solely transferring demonstrations across robot embodiments~\cite{lepert2024shadow,chen2024mirage,chen2024roviaug}, without combining these into a unified pipeline.

To address these gaps, we present a novel and \textit{unified} data augmentation framework for visual imitation learning, called \textbf{\underline{C}}anny-guided \textbf{\underline{R}}obot D\textbf{\underline{a}}ta Generation using Video Di\textbf{\underline{f}}fusion \textbf{\underline{T}}ransformers (\method). \method constructs a digital twin of a real-world setup to generate simulation trajectories, extracts Canny-edge control videos, and conditions a video diffusion model~\cite{wan2025} on these edges alongside a real-world reference image and language instruction. Our key insight is that Canny-edge conditioning guides the diffusion model by preserving the most relevant structural contours while abstracting away low-level simulation details. This enables a unified augmentation pipeline spanning object pose, object color, background, lighting, camera viewpoints, wrist and third-person multi-view generation, and bimanual cross-embodiment transfer.

We evaluate \method across three simulated and three real-world bimanual manipulation tasks using the RoboTwin~\cite{chen2025robotwin} benchmark and a physical setup with two bimanual xArm7 and Franka arms. See Figure~\ref{fig:teaser} for a visualization of the diverse synthetic training videos produced by \method, spanning seven axes of visual variation. Using ACT~\cite{Zhao-RSS-23} as our downstream policy, \method consistently improves robustness and generalization over existing augmentation baselines across all seven augmentation techniques.

The contributions of this paper include:
\begin{itemize}
    \item \method, a novel and unified method to utilize Canny edge images~\cite{canny1986edge} as a control input to condition video generative models to generate high-quality and diverse robot videos. 
    \item A new pipeline for bimanual cross-embodiment manipulation that can perform additional image augmentations compared to prior approaches. %
    \item Simulation and real-world experiments demonstrating that policies trained on \method-generated data significantly outperform baselines, with ablations quantifying the benefit of each augmentation technique.
\end{itemize}

\section{Related Work}

\subsection{Video Generation For Robotics}

Recent years have seen the emergence of powerful diffusion-based~\cite{ho2020denoisingdiffusionprobabilisticmodels} video generative models that can generate high-fidelity video frames given conditioning inputs (e.g., text or images)~\cite{nvidia2025cosmosworldfoundationmodel,yang2026cogvideox}. 
In this work, we use Wan 2.1~\cite{wan2025} but other video generative models are applicable if they can condition on modalities such as Canny edge~\cite{canny1986edge} images.
Some works use video prediction models to generate robot trajectories~\cite{jang2025dreamgenunlockinggeneralizationrobot} or action-conditioned predictions with wrist and third-person views~\cite{guo2026ctrlworldcontrollablegenerativeworld}, while others reconstruct physically grounded world models for policy training~\cite{mao2026robotlearningphysicalworld}.
In contrast, we use video diffusion to synthesize additional action-labeled demonstrations for imitation learning using a simulator, without learning a predictive world model.
More relevant to our work, AnchorDream~\cite{ye2026anchordream} conditions generation on rendered robot motion traces without explicit simulator rollouts, but the lack of a simulator limits augmentation diversity. \method instead leverages a simulator and digital twin pipeline to produce physically plausible trajectory variations across diverse scene configurations, yielding more diverse, higher-fidelity demonstrations across bimanual and multi-view settings, though at the cost of requiring access to a simulator and object meshes.

\subsection{Cross-Embodiment Learning}

A growing body of work tackles data scarcity in robotics by editing or synthesizing images to reduce distribution shift when transferring to a target robot. 
Mirage~\cite{chen2024mirage} performs ``cross-painting'' by inpainting the target robot over a source robot at evaluation time, enabling zero-shot policy transfer. 
RoVi-Aug~\cite{chen2024roviaug} extends this idea by leveraging diffusion models to generate images of a target robot to expand training data. 
Shadow~\cite{lepert2024shadow}, instead, aligns train and test distributions through deterministic and composite robot mask overlays, and avoids pixel-level inpainting. 
While effective for single-arm cross-embodiment data transfer, these methods have not been evaluated for bimanual manipulation. Moreover, they do not generate new action labels, in contrast to our proposed framework. 
Another direction in cross-embodiment learning trains a policy on mixed data from multiple robot embodiments~\cite{yang2024pushinglimits,doshi2024scalingcrossembodied,bauer2025latentactiondiffusioncrossembodiment}, but this requires extensive data from each robot embodiment. 
Finally, other works in cross-embodiment learning study applications in orthogonal domains such as inverse reinforcement learning~\cite{zakka2021xirl} and human-robot cross-embodiment~\cite{lum2025crossinghumanrobotembodimentgap,dan2025xsim}, which are beyond the scope of our work.

\subsection{Data Augmentation for Imitation Learning}

Data augmentation has emerged as a practical tool for scaling imitation learning without additional human demonstrations. 
Prior work uses generative models to alter visual context such as backgrounds or objects while keeping actions fixed~\cite{chen2024semanticallycontrollable,yuan2025roboengineplugandplayrobotdata,bharadhwaj2024roboagent,yu2023scaling,wang2026roboaug}, in contrast to state-based augmentation approaches~\cite{mitrano2022dataaugmentationmanipulation,ke2024ccil,Laskey2017DARTNI}. 
\method similarly adjusts visual context but additionally expands the action distribution and does not require high-fidelity scene reconstruction as in~\cite{robosplat}. 
Recent work has also studied viewpoint augmentation, from third-person perspectives for single-arm~\cite{tian2024vista} and bimanual~\cite{chen2026ropasyntheticrobotpose} contexts, as well as from wrist-cameras~\cite{liu2025DCODA,zhou2023nerfpalmhand,zhang2024diffusionmeetsdagger,ding2025imaginationinferencesynthesizinginhand,xie2026multicameraviewscalingdataefficient}, but none unify both views in a single framework which \method supports.
Finally, another direction explores both state and action data augmentation through Real2Sim2Real simulation rollouts~\cite{mandlekar2023mimicgen,jiang2025dexmimicen}. While our method also uses a digital twin, it does not require the last Sim2Real step, since we employ a video diffusion model to synthesize photorealistic images from simulator-generated trajectories.  
This allows us to expand trajectory diversity in a way that preserves coordination constraints and contact dynamics to facilitate data-efficient bimanual manipulation. 

\begin{figure*}[t]
    \centering
\includegraphics[width=1.0\textwidth]{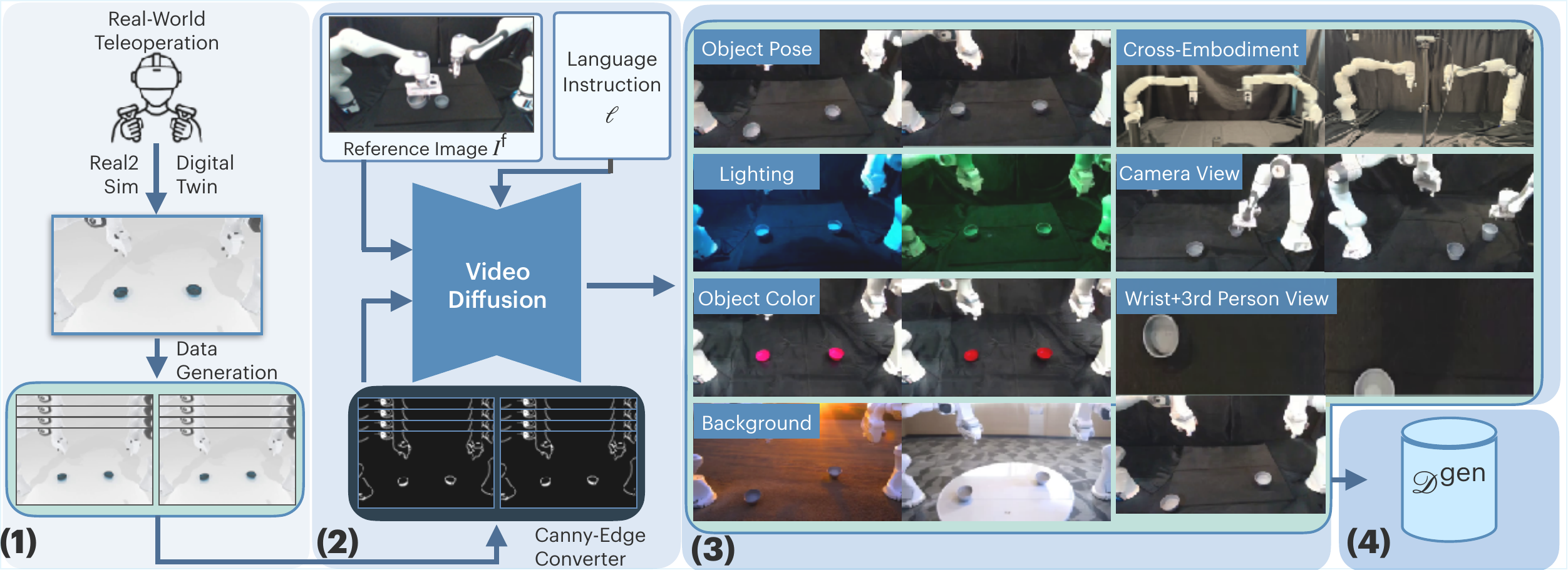}
    \caption{
        \textbf{Method Overview.} \textbf{(1) Trajectory Expansion:} Real-world teleoperation data is first collected, and a digital twin pipeline transfers the objects and robot into simulation (Real2Sim). This simulation environment is then used for large-scale data generation. \textbf{(2) Video Generation:} The simulation trajectories are rendered into source videos and passed through a Canny-Edge Converter to extract structural edge representations, which are then combined with a real-world reference image and language instructions to condition a video diffusion model that synthesizes photorealistic video outputs. \textbf{(3) Augmented Dataset Construction:} The resulting generated videos support a wide range of visual variations, including object pose, lighting conditions, object color, background, cross-embodiment transfer, camera viewpoint, and combined wrist and third-person camera perspectives. \textbf{(4) Generated Dataset:} The synthesized videos are paired with action labels from the simulation trajectories, producing action-consistent demonstrations $\mathcal{D}^{\text{gen}}$ for downstream policy training.
    }
    \vspace{-10pt}
    \label{fig:pipeline}
\end{figure*}

\section{Problem Statement}
\label{sec:ps}
Our focus is on scalable data generation for vision-based imitation learning in bimanual manipulation, where a policy $\pi_\theta$ with parameters $\theta$ is trained from expert demonstrations using third-person RGB, wrist-camera, or combined RGB image observations. We denote a camera image at time $t$ as $I_t$, simulation-generated images as $I^{\source}_{t}$, video-diffusion-synthesized images as $I^{\generated}_{t}$, and ground truth deployment images as $I^{\gt}_{t}$. At deployment, the policy receives $I^{\text{g}}_{t}$ and produces actions $a_t = \pi_\theta(I^{\text{g}}_{t})$, where $a_t = (a^{\lef}_{t}, a^{\rig}_{t})$ specifies target joint positions and gripper actuation for the left and right arms, respectively.

We assume access to a small set of $M$ real-world teleoperation demonstrations $\mathcal{D}^{\text{real}} = \{\tau^{\text{real}}_1, \ldots, \tau^{\text{real}}_M\}$ and a simulation environment generating source videos $\mathbf{V}^{\source}$ via a digital twin pipeline. Each demonstration %
is a sequence of ground truth image observations and corresponding actions:
\begin{equation}
    \tau^{\text{real}}_i = (I^{\gt}_{1}, a^{\lef}_{1}, a^{\rig}_{1}, \ldots, I^{\gt}_{T}, a^{\lef}_{T}, a^{\rig}_{T}),
\end{equation}
for a demonstration of $T$ timesteps. Our goal is to synthesize a large, visually diverse set of generated demonstrations $\mathcal{D}^{\text{gen}}$, with $|\mathcal{D}^{\text{gen}}| \gg |\mathcal{D}^{\text{real}}|$, where each synthesized demonstration contains diffusion-synthesized observations $I^{\generated}_{t}$ resembling real-world images, to train a policy on $\mathcal{D}^{\text{real}} \cup \mathcal{D}^{\text{gen}}$.  %

\section{Method: \method}

\method leverages a video diffusion model to synthesize photorealistic and visually diverse training videos for bimanual manipulation. Given a simulation-generated source video $\mathbf{V}^{\source}$ produced from a digital twin pipeline, a real-world reference image $I^{\text{f}}$, and a language instruction $\ell$, the model outputs a photorealistic target video $\mathbf{V}^{\generated} = \{I^{\generated}_{1}, \ldots, I^{\generated}_{T}\}$ that preserves robot motion structure while matching diverse real-world visual appearance. This is achieved through three stages: trajectory expansion (Section~\ref{ssec:trajectoryexpansion}), video generation (Section~\ref{ssec:video-generation}), and augmented dataset construction for policy training (Section~\ref{ssec:dataset-construction}). 
\method repeatedly applies this procedure to obtain $\mathcal{D}^\text{gen}$. 
Figure~\ref{fig:pipeline} provides an overview. 

\subsection{Trajectory Expansion}
\label{ssec:trajectoryexpansion}

We construct a simulation counterpart $\mathcal{D}^{\text{sim}}$ from $\mathcal{D}^{\text{real}}$ using a digital twin pipeline, leveraging AprilTags~\cite{wang2016apriltag} for object localization and known object meshes from RoboTwin~\cite{chen2025robotwin}, though any pipeline reconstructing object meshes and robot models in simulation is applicable~\cite{chen2024urdformer}. Each real trajectory $\tau_i^\text{real}$ is replayed in simulation to generate a source video $\mathbf{V}^\source$ and corresponding simulation trajectory $\tau_i^\text{sim}$, resulting in a new simulation data of equal size as the original: $|\mathcal{D}^\text{sim}| = |\mathcal{D}^\text{real}|$.

\begin{figure*}[t]
    \centering
    \includegraphics[width=1.0\textwidth]{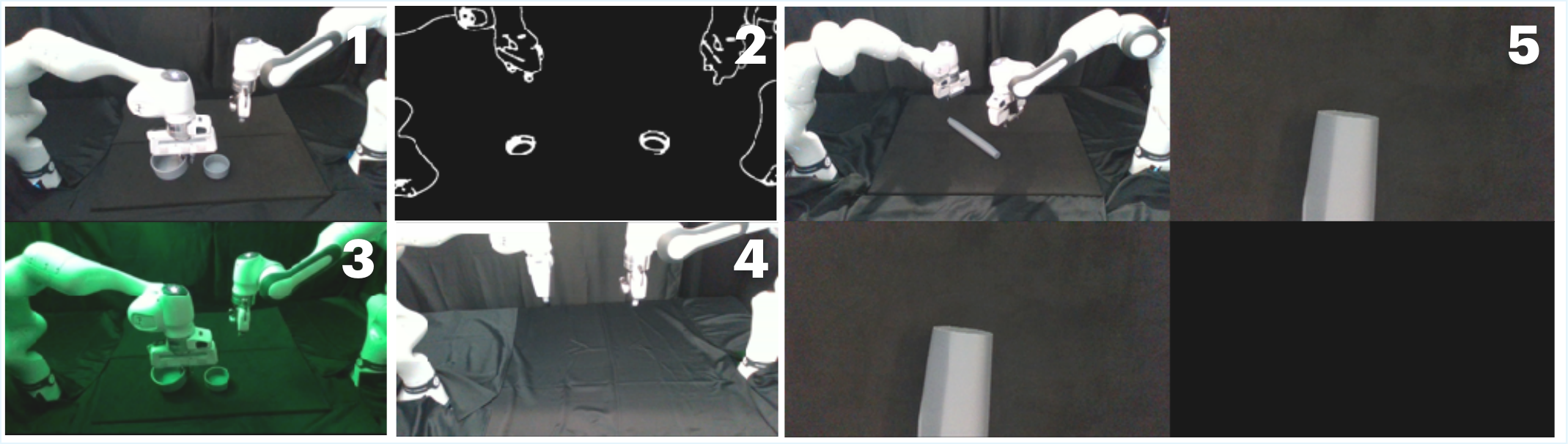}
    \caption{
        \textbf{Reference Image and Canny-edge Visualization.} Examples of reference images used to condition the video diffusion model for different augmentation techniques: \textbf{(1)} A standard reference image of the scene capturing gripper-object contact, used to condition the video diffusion model. \textbf{(2)} An example \textbf{Canny-edge frame} extracted from the simulation source video $\mathbf{V}^{\text{c}}$, used as structural control input. \textbf{(3)} A lighting-modified reference image generated using Veo3~\cite{google2025veo3} under green ambient illumination. \textbf{(4)} An empty table reference image with no objects, used for object color generation. \textbf{(5)} A tiled reference image combining a third-person view (top left), left wrist (top right), and right wrist (bottom left), with the fourth tile left blank, supporting up to four simultaneous camera viewpoints. Reference images include top and bottom padding (not shown).
    }
    \label{fig:reference_images}
    \vspace{-10pt}
\end{figure*}

To scale up data collection, we expand $\mathcal{D}^{\text{sim}}$ inspired by DexMimicGen~\cite{jiang2025dexmimicen}: each trajectory $\tau_i^{\text{real}}$ is decomposed into object-centric subtasks by annotating per-arm timestep boundaries, and a transformation operator $\mathcal{T}$ is applied to produce a new candidate trajectory $\mathcal{T}(\tau^{\text{real}}_i)$ consistent with a novel sampled scene configuration.
Each candidate is executed in simulation and validated for task success, retaining only successful trajectories to expand $\mathcal{D}^{\text{sim}}$. 
The validated trajectories are rendered into source videos $\mathbf{V}^{\source}$, from which Canny-edge control videos $\mathbf{V}^{\text{c}}$ are extracted by filtering for salient structural edges (Figure~\ref{fig:reference_images}, Tile 2), and fed into the video generation stage (Section~\ref{ssec:video-generation}) to synthesize $\mathcal{D}^{\text{gen}}$.

\subsection{Video Generation}
\label{ssec:video-generation}
Diffusion-based video generative models learn to approximate a distribution over video sequences through iterative denoising. The model learns the conditional distribution:
\begin{equation}
p_\phi(\mathbf{V}^{\generated} \mid I^{\text{f}}, \mathbf{V}^{\text{c}}, \ell),
\end{equation} where $\mathbf{V}^{\text{c}}$ is the Canny-edge control video from Section~\ref{ssec:trajectoryexpansion}, $I^{\text{ref}}$ is the real-world reference image, $\ell$ is the language instruction, and $\phi$ denotes the model parameters. We use Wan2.1-Fun-Control~\cite{wan2025} as our backbone, which supports Canny-edge, depth, and skeleton pose conditioning. We choose Canny-edge conditioning over depth or skeleton pose because skeleton pose captures only robot arm structure without encoding object information, while depth conditioning retains too much scene detail, reducing the model's flexibility to synthesize diverse visual appearances. Canny edges balance both by preserving salient structural features of robot arms and objects while discarding fine-grained details, giving the model freedom to vary visual appearance. By selectively varying $I^{\text{f}}$ and $\ell$ while keeping $\mathbf{V}^{\text{c}}$ fixed, the model synthesizes diverse target videos $\mathbf{V}^{d}$ without altering robot motion structure. Depending on the desired variation, we modify $I^{\text{f}}$, $\ell$, or both, and leverage an LLM to automatically generate semantic variants of $\ell$ (see Section~\ref{ssec:dataset-construction}).

\subsection{Augmented Dataset Construction}
\label{ssec:dataset-construction}
We leverage Canny edges to guide demonstration augmentation across seven dimensions: object pose, lighting, object color, background, cross-embodiment, camera viewpoint, and wrist with third-person view generation. \textit{\method is flexible and modular where users can apply any subset of augmentation techniques and control the number of generated demonstrations.} Several augmentation techniques leverage LLMs to automatically generate diverse prompts and complete prompt lists are provided in the supplementary material.

\subsubsection{Object Pose}
\label{ssec:object-pose}
To augment object poses, we introduce variations during trajectory expansion (Section~\ref{ssec:trajectoryexpansion}). For each source trajectory $\tau_i^\text{real}$, the simulator applies random translations and rotations to the target object's pose, sampled from a uniform distribution with ranges set based on the physically feasible workspace. We also find that using a reference image capturing gripper-object contact yields higher fidelity contact synthesis in generated videos (Figure~\ref{fig:reference_images}, Tile 1).

\subsubsection{Lighting}
To generate diverse lighting conditions, we augment the reference image $I^{\text{f}}$ by prompting an image generation model, Veo3~\cite{google2025veo3}, to synthesize variants under different ambient illumination, such as blue or green lighting. Unlike simple color jitter or RGB channel manipulation, this approach preserves scene properties such as shadows and surface reflections (Figure~\ref{fig:reference_images}, Tile 3). The augmented reference images are then used to condition the video diffusion model, producing target videos $\mathbf{V}^{\generated}$ with photorealistic lighting variations while preserving the underlying robot motion structure.

\subsubsection{Object Color}
To generate diverse object colors, we use a reference image $I^{\text{f}}$ of the empty table scene without any objects as illustrated in Figure~\ref{fig:reference_images} (Tile 4). Conditioning on a reference image that contains objects would anchor the generated scene to the object color present in the reference, limiting color diversity. Since the reference image contains no objects, the Canny-edge control video $\mathbf{V}^{\text{c}}$ provides the object contours to inform the diffusion model of their location, while the language instruction $\ell$ specifies the desired color to guide the appearance of the synthesized objects. By modifying the language instruction to specify the desired object color, the video diffusion model synthesizes target videos $\mathbf{V}^{\generated}$ with the specified object appearance while preserving the scene layout and robot motion structure. To avoid manual prompt editing, we prompt an LLM to generate a list of object colors, from which we sample randomly during dataset construction.

\subsubsection{Background}
To generate diverse backgrounds, we omit the reference image $I^{\text{f}}$ from the video diffusion model, as conditioning on it anchors the generated scene to the original environment. Instead, we modify the instruction $\ell$ to describe the desired background. To scale background diversity without manual prompting, we leverage an LLM to automatically generate a large set of varied background descriptions, which are then used to condition the video diffusion model to produce target videos $\mathbf{V}^{\generated}$ with diverse scene appearances. 

\subsubsection{Cross-Embodiment}
To enable cross-embodiment transfer, we map demonstrations from a source robot to a target robot using forward and inverse kinematics, and replace images of the source robot with photorealistic images of the target robot. This allows us to directly use the transferred demonstrations as training data for the target robot, without requiring any additional real-world data collection. At each timestep $t$, we apply forward kinematics $\text{FK}(\cdot)$ to extract the end-effector poses for the left and right arms:
\begin{equation}
    p^{\lef}_{t} = \text{FK}(a^{\lef}_{t}), \quad p^{\rig}_{t} = \text{FK}(a^{\rig}_{t}),
\end{equation}
where $p^{\lef}_{t}$ and $p^{\rig}_{t}$ denote the Cartesian end-effector positions for the left and right arms, respectively. We then apply inverse kinematics $\text{IK}(\cdot)$ to compute the corresponding joint configurations for the target robot:
\begin{equation}
    \hat{a}^{\lef}_{t} = \text{IK}(p^{\lef}_{t}), \quad \hat{a}^{\rig}_{t} = \text{IK}(p^{\rig}_{t}),
\end{equation}
while preserving the gripper actuation from the source trajectory $\tau_i^\text{real}$. The retargeted demonstrations are replayed in simulation to generate the Canny-edge control video $\mathbf{V}^{\text{c}}$, which conditions the video diffusion model to synthesize target videos $\mathbf{V}^{\generated} = \{I^{\generated}_{1}, \ldots, I^{\generated}_{T}\}$ of the new robot embodiment.

\subsubsection{Camera Viewpoint}
To generate diverse camera viewpoints, we place additional cameras inside the simulator and tile up to four simultaneous views into a single image. Formally, given $1 \le N \leq 4$ camera views $\{I^{\source,1}_{t}, \ldots, I^{\source,N}_{t}\}$, we construct a tiled source image $I^{\source,\text{tile}}_{t} = \{I^{\source,1}_{t}, \ldots, I^{\source,N}_{t}\}$
from which the Canny-edge control video $\mathbf{V}^{\text{c}}$ is extracted. The reference image $I^{\text{f}}$ is similarly tiled to match, and both are fed into the video diffusion model to synthesize target videos $\mathbf{V}^{\generated}$ across multiple camera perspectives simultaneously. The video diffusion model automatically preserves the tiled structure in the generated output and each viewpoint remains spatially contained within its corresponding tile without going into adjacent tiles. Figure~\ref{fig:reference_images} (Tile 5) shows an example of the tiled reference image using wrist cameras; for camera view augmentation, the wrist camera tiles would instead be replaced with the desired third-person camera viewpoints.

\subsubsection{Wrist and Third-Person View}
Here, we follow the same tiling approach as camera viewpoint generation. Instead of tiling multiple third-person views, we tile the left wrist camera $I^{\source,\lef}_{t}$, right wrist camera $I^{\source,\rig}_{t}$, and a third-person (external) camera $I^{\source,\text{ext}}_{t}$ into a single image $I^{\source,\text{tile}}_{t} = \{I^{\source,\lef}_{t}, I^{\source,\rig}_{t}, I^{\source,\text{ext}}_{t}, \emptyset\}$, leaving the fourth tile empty, from which the Canny-edge control video $\mathbf{V}^{\text{c}}$ is extracted. Tiling ensures spatial consistency across all viewpoints, and the reference image $I^{\text{f}}$ is tiled accordingly as shown in Figure~\ref{fig:reference_images} (Tile 5) before being fed into the video diffusion model to synthesize target videos $\mathbf{V}^{\generated}$.

\begin{figure}[t]
    \centering
    \includegraphics[width=0.48\textwidth]{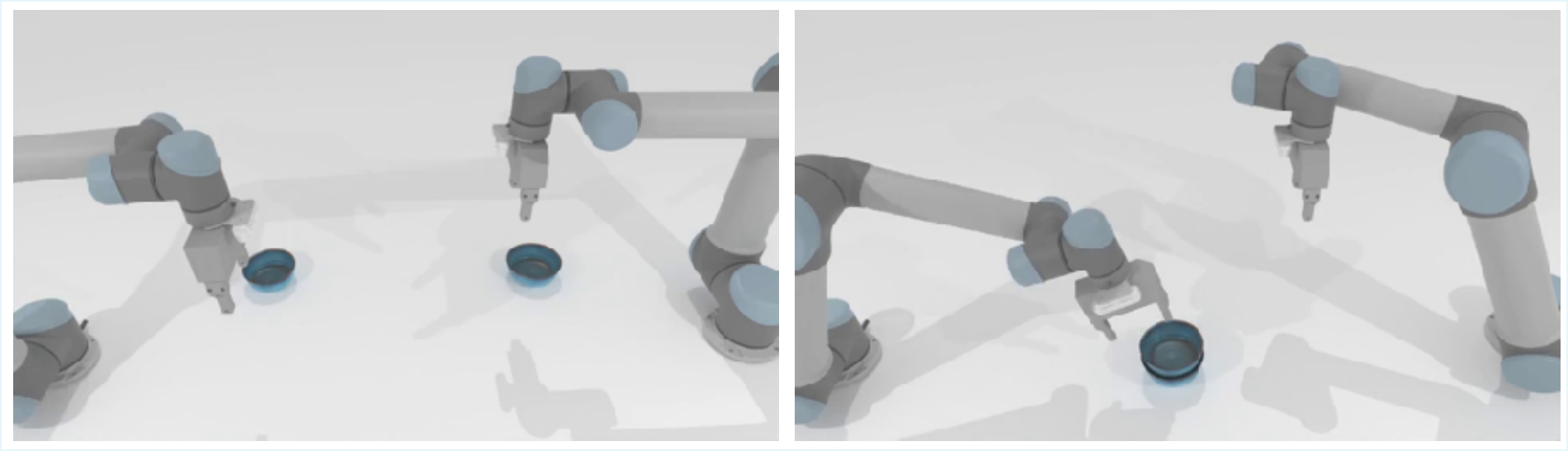}
    \caption{
        \textbf{Simulation Environment.} The \texttt{SB} (see Section~\ref{sec:sim_exps}) task adapted from RoboTwin~\cite{chen2025robotwin} on the bimanual UR5 with WSG grippers, showing the initial state (left) and success state (right).
    }
    \label{fig:simulation}
    \vspace{-10pt}
\end{figure}

\section{Simulation Experiments}
\label{sec:sim_exps}

Simulation experiments are conducted to evaluate the cross-embodiment capabilities of \method, using the open-source RoboTwin~\cite{chen2025robotwin} benchmark for bimanual manipulation.
We modify the tasks to better align with the Action Chunking with Transformers (ACT)~\cite{Zhao-RSS-23} policy and evaluate cross-embodiment transfer from a bimanual UR5 with WSG grippers (see Figure~\ref{fig:simulation}) to a bimanual Franka Panda across three tasks spanning different coordination strategies:

\begin{itemize}
    \item \texttt{\textbf{Lift Pot (LP):}} A coordinated bimanual task where both arms must simultaneously grasp and lift a pot.
    \item \texttt{\textbf{Place Cans in Plasticbox (PC):}} A parallel task where both arms independently pick up cans and place them into a container.
    \item \texttt{\textbf{Stack Bowls (SB):}} A sequential task where two bowls must be stacked on top of each other in order.
\end{itemize}

We evaluate the cross-embodiment variant of \method, where the goal is to transfer demonstrations collected on a bimanual UR5 with WSG grippers (source robot) to train a policy for a bimanual Franka Panda (target robot) in simulation, without collecting any demonstrations on the target robot.

\begin{itemize}
    \item \textbf{Collected Target:} Demonstrations collected directly on the target robot, serving as an upper-bound reference for cross-embodiment performance.
    \item \textbf{Shadow~\cite{lepert2024shadow}:} A data editing approach that replaces robot observations with a composite segmentation mask of the source and target robots, which we extend to bimanual settings by masking both arms simultaneously.
    \item \textbf{\method~(Target):} Our method applied using Wan2.1-Fun-Control, without any additional generated demonstrations beyond the collected robot data.
    \item \textbf{\method~(Ours):} Our full pipeline, which expands the source robot dataset with 1000 generated demonstrations using object pose, lighting, and background augmentation before transferring to the target robot embodiment.
\end{itemize}

To ensure fair comparisons, all methods use identical training, validation, and test splits with consistent environment seeds, evaluated over 3 random seeds. Video generation is performed \textbf{zero-shot} using the Wan2.1-Fun-Control 1.3B model. ACT policy training and inference are conducted on a single NVIDIA RTX 4090 GPU. For small-scale video generation we use a single NVIDIA RTX 5090, while large-scale generation is distributed across 3 NVIDIA RTX 6000 GPUs. At inference time, input images are center-cropped, padded, and resized using OpenCV to $512 \times 512$ pixels before being passed to the video generation model.

\begin{table}
  \centering
  \resizebox{\columnwidth}{!}{%
  \begin{tabular}{lcccccc}
    \toprule
    & \multicolumn{3}{c}{\textbf{Simulation (\%)}} & \multicolumn{3}{c}{\textbf{Real-World ( / 20 )}} \\
    \cmidrule(lr){2-4} \cmidrule(lr){5-7}
    Method & LP & PC & SB & LR & PC & SB \\
    \midrule
    Collected Target & 55.0\% & 69.0\% & 59.0\% & 5 / 20 & 2 / 20 & 3 / 20 \\
    \midrule
    Shadow~\cite{lepert2024shadow} & 2.0\% & 2.3\% & 6.0\% & 2 / 20 & 1 / 20 & 1 / 20 \\
    \method~(Target) & 11.3\% & 6.0\% & 21.6\% & 4 / 20 & 1 / 20 & 2 / 20 \\
    \rowcolor{lightgreen} \method~(Ours) & \textbf{82.6\%} & \textbf{89.3\%} & \textbf{86.0\%} & \textbf{17 / 20} & \textbf{15 / 20} & \textbf{16 / 20} \\
    \bottomrule
  \end{tabular}%
  }
  \caption{
   \textbf{Cross-Embodiment Results.} Simulation success rates (\%) and real-world successes out of 20 trials for cross-embodiment transfer. Simulation evaluates transfer from a bimanual UR5 to a bimanual Franka Panda on \texttt{LP}, \texttt{PC}, and \texttt{SB} (see Section~\ref{ssec:sim_results}). Real-world evaluates transfer from a bimanual xArm7 to a bimanual Franka Panda on \texttt{LR}, \texttt{PC}, and \texttt{SB} (see Section~\ref{ssec:real-world-results}). \method~(Target) denotes source-to-target transfer demos, with the same number of demos as Target. \method~(Ours) uses 1000 generated demos without collecting any target robot demos, in contrast to Collected Target which requires 100 demos on the target robot. All methods are trained and evaluated using ACT.
    }
    \vspace{-6pt}
  \label{tab:cross-embodiment}
\end{table}

\begin{table}
  \centering
  \begin{tabular}{lc}
    \toprule
    \textbf{Method} & \textbf{Stack Bowls (SB)} \\
    \midrule
    Collected Demos (Upper Bound) & 59.0\% \\
    \midrule
    \method~w/o Canny &  10.3\%\\
    \rowcolor{lightgreen} \method~w/ Canny &  21.6\% \\
    \bottomrule
  \end{tabular}%
  \caption{
    \textbf{Ablation Study.} Success rates out of 150 demonstrations on \texttt{Stack Bowls} comparing video generation with and without Canny-edge control input. Collected Demos (Upper Bound) serves as the upper bound. No augmentation is performed in this ablation, as the objective is to isolate and evaluate the effect of synthesized image quality on ACT performance.
  }
  \vspace{-10pt}
  \label{tab:ablation}
\end{table}

\subsection{Simulation Results}
\label{ssec:sim_results}
As shown in Table~\ref{tab:cross-embodiment}, our Target variant already outperforms Shadow, which struggles on precision-demanding tasks as its segmentation mask occludes critical gripper-object contact points, degrading ACT performance. 
While the Target variant better preserves visibility, it alone does not yield substantial improvement. 
Scaling to 1000 generated demonstrations leads to substantially higher success rates: $11.3\%\to 82.6\%$, $6.0\%\to 89.3\%$, and $21.6\%\to 86.0\%$.
Notably, \method~(Ours) surpasses the target demonstration baseline (target robot demos) despite never collecting target robot data, showing that diverse data generation spanning varied object poses is a scalable alternative to target robot data collection. 

\subsection{Ablation Studies}
\label{sec:ablation}

We examine whether converting simulation videos to Canny-edge representations improves generation quality over raw simulation images.  
We replace $\mathbf{V}^{\text{c}}$ with $\mathbf{V}^{\source}$ as input, referring to this variant as \textbf{\method~w/o Canny}. 
Note that no additional data augmentation is applied in this ablation since we solely isolate the effect of Canny-edge conditioning on the quality of synthesized images, comparing generated demonstrations against the collected demonstration reference. 
As shown in Table~\ref{tab:ablation}, \method~w/ Canny achieves nearly twice the success rate of \method~w/o Canny on \texttt{Stack Bowls}. 
This is because raw simulation images retain too much low-level detail, causing the diffusion model to struggle with salient structural features such as gripper-object contact, leading to degraded synthesis. Canny edges discard irrelevant details while preserving robot arm and object structure, giving the diffusion model clear guidance on what to generate. 

\begin{table*}
\centering
\resizebox{\textwidth}{!}{%
\begin{tabular}{@{}lcccccccccccccccc@{}}
\toprule
& \multicolumn{3}{c}{\textbf{Lighting}} & \multicolumn{3}{c}{\textbf{Background}} & \multicolumn{3}{c}{\textbf{Camera View}} & \multicolumn{3}{c}{\textbf{Object Color}} & \multicolumn{3}{c}{\textbf{Wrist + 3rd Person}} \\
\cmidrule(lr){2-4} \cmidrule(lr){5-7} \cmidrule(lr){8-10} \cmidrule(lr){11-13} \cmidrule(lr){14-16}
\textbf{Method} & LR & PC & SB & LR & PC & SB & LR & PC & SB & LR & PC & SB & LR & PC & SB \\
\midrule
ACT w/o Aug & 3 / 20 & 1 / 20 & 0 / 20 & 4 / 20 & 0 / 20 & 0 / 20 & 6 / 20 & 3 / 20 & 2 / 20 & 2 / 20 & 0 / 20 & 1 / 20 & 15 / 20 & 11 / 20 & 13 / 20 \\
\method Pose-Only & 5 / 20 & 3 / 20 & 2 / 20 & 7 / 20 & 2 / 20 & 3 / 20 & 13 / 20 & 5 / 20 & 7 / 20 & 5 / 20 & 2 / 20 & 3 / 20 & 13 / 20 & 8 / 20 & 10 / 20 \\
ACT w/ Baseline Aug & 13 / 20 & 9 / 20 & 8 / 20 & 4 / 20 & 5 / 20 & 6 / 20 & 14 / 20 & 8 / 20 & 6 / 20 & 15 / 20 & 9 / 20 & 11 / 20 & N/A$^\dagger$ & N/A$^\dagger$ & N/A$^\dagger$ \\
\rowcolor{lightgreen}\method (Ours) & \textbf{17 / 20} & \textbf{14 / 20} & \textbf{12 / 20} & \textbf{18 / 20} & \textbf{15 / 20} & \textbf{10 / 20} & \textbf{19 / 20} & \textbf{18 / 20} & \textbf{18 / 20} & \textbf{18 / 20} & \textbf{18 / 20} & \textbf{17 / 20} & \textbf{20 / 20} & \textbf{19 / 20} & \textbf{20 / 20} \\
\bottomrule
\end{tabular}%
}
\begin{minipage}{\textwidth}
  \footnotesize $^\dagger$ No suitable baseline augmentation method exists for this augmentation type.
\end{minipage}
\caption{\textbf{Real-World Augmentation Results.} Success rates out of 20 for \texttt{LR}, \texttt{PC}, and \texttt{SB} (see Section~\ref{ssec:real-world-results}) across five augmentation techniques. For each augmentation technique, all methods are evaluated under test conditions that vary only along that specific dimension (e.g., unseen lighting conditions for Lighting, unseen backgrounds for Background), while all other visual factors remain fixed. All \method~(Ours) columns use 1000 generated demonstrations combined with the real-world collected demonstrations (100 for \texttt{LR}, 200 for \texttt{PC}, and 150 for \texttt{SB}), where the number of collected demonstrations per task was determined based on the minimum required to achieve reasonable ACT policy performance. Baseline methods use the same number of collected demonstrations augmented with their respective augmentation technique. The ``ACT w/ Baseline Aug'' row refers to a \emph{different} baseline method for each augmentation type, chosen as the most relevant prior work for a specific augmentation: Lighting (Color Jitter), Background (RoboEngine~\cite{yuan2025roboengineplugandplayrobotdata}), Camera View (VISTA~\cite{tian2024vista}), and Object Color (SAM3~\cite{carion2025sam3segmentconcepts}). All methods are trained and evaluated using an ACT policy on the bimanual Franka.}
\label{tab:augmentation_comparison}
\vspace{-10pt}
\end{table*}

\begin{figure}[t]
    \centering
    \includegraphics[width=0.48\textwidth]{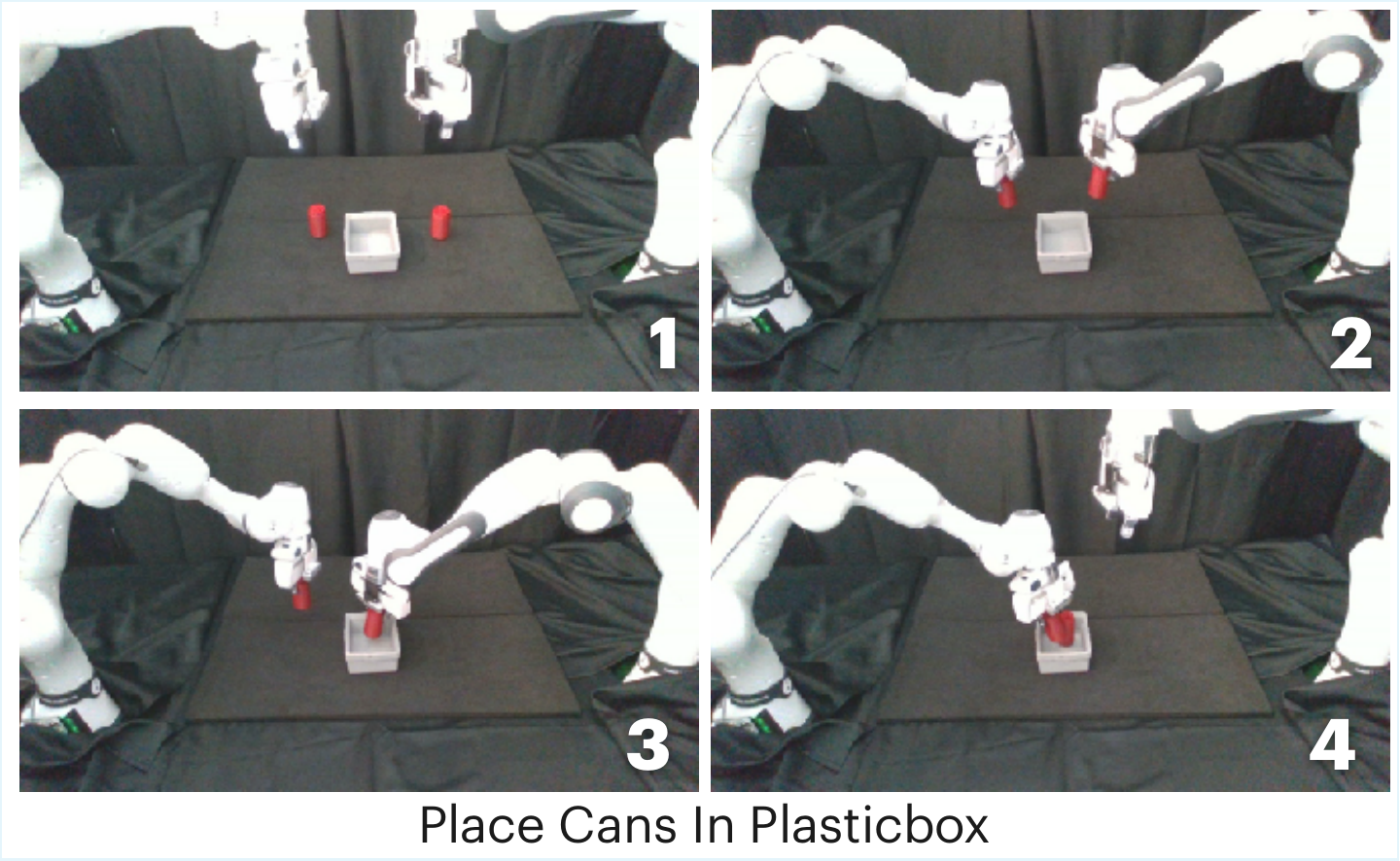}
    \caption{
    \textbf{Real-world rollout} trained on ACT of the \texttt{PC} (see Section~\ref{ssec:real-world-results}), with the bottom-right corner of each image showing the progression of the task execution. Images are labeled in order of task progression.
}
    \label{fig:real-world-rollouts}
    \vspace{-10pt}
\end{figure}

\section{Real-World Experiments}

\subsection{Real-World Experiment Setup}
We use a bimanual Franka Research 3 with GELLO~\cite{wu2024gello}, one or three Intel RealSense D435i cameras depending on whether wrist-camera observations are needed, an NVIDIA RTX 5090 for ACT training and inference, and \textbf{zero-shot} video generation via Wan2.1-Fun-Control 1.3B.
AprilTags from the third-person camera estimate object positions and camera extrinsics, with object meshes sourced from RoboTwin, though other digital twin pipelines~\cite{chen2024urdformer} can be substituted.
We evaluate each policy's success rate over 20 trials per task across three tasks spanning a range of bimanual coordination strategies. 

\begin{itemize}
    \item \texttt{\textbf{Lift Roller (LR):}} A coordinated task where both arms simultaneously grasp and lift a dough roller.
    \item \texttt{\textbf{Place Cans in Plasticbox (PC):}} A parallel task where both arms independently pick up cans and place them into a container.
    \item \texttt{\textbf{Stack Bowls (SB):}} A sequential task where two bowls must be stacked on top of each other in order.
\end{itemize}

\subsection{Real-World Results}
\label{ssec:real-world-results}
We evaluate \method across seven augmentation techniques (see Section~\ref{ssec:dataset-construction}): object pose, lighting, object color, background, cross-embodiment, camera viewpoint, and wrist with third-person view generation. For each augmentation type, we compare policies trained on: (1) real-world demonstrations only (i.e., $\mathcal{D}^\text{real}$), (2) demonstrations with object pose augmentation only (\method~Pose-Only), (3) real-world demonstrations augmented with an augmentation-specific baseline method, and (4) demonstrations generated via our full \method pipeline. \method~Pose-Only is included to isolate the contribution of object pose variation from the remaining augmentation techniques.

For each augmentation type, we compare the same baselines and evaluate under test conditions that vary only along that specific dimension. For example, unseen lighting conditions for lighting and unseen backgrounds for background, while all other visual factors remain fixed. We provide further details on the test conditions for each augmentation techniques in the supplementary material. Due to task simplicity, \texttt{Lift Roller} uses fewer demonstrations than \texttt{Place Cans in Plasticbox} and \texttt{Stack Bowls} across all methods:
\begin{itemize}
    \item \textbf{ACT w/o Aug:} 50 (\texttt{LR}) / 100 (\texttt{PC}, \texttt{SB}) real-world demonstrations collected under standard conditions trained on ACT~\cite{Zhao-RSS-23}.
    \item \textbf{\method~Pose-Only:} 100 (\texttt{LR}) / 200 (\texttt{PC}, \texttt{SB}) demonstrations with object pose augmentation only. Inspired by RoboSplat~\cite{robosplat}, we include this baseline to assess the standalone impact of varying object poses. %
    \item \textbf{ACT with Baseline Aug (augmentation-specific):} 50 (\texttt{LR}) / 100 (\texttt{PC}, \texttt{SB}) demonstrations with an augmentation-specific method. The specific baseline used for each augmentation type is noted in the corresponding subsection.
    \item \textbf{\method~(Ours):} 1000 (\texttt{LR}, \texttt{PC}, \texttt{SB}) generated demonstrations and the original real-world demonstrations using our full augmentation pipeline.
\end{itemize}

\subsubsection{Lighting}
To evaluate lighting generalization, we test policy deployment under four distinct lighting conditions: blue, green, red, and yellow ambient illumination. We use \textbf{Color Jitter} as the augmentation-specific baseline, implemented by applying random per-channel scale offsets and Gaussian noise to each frame using OpenCV; we adopt this baseline as it was used in RoboSplat~\cite{robosplat} for the same purpose of evaluating lighting robustness. As shown in Table~\ref{tab:augmentation_comparison}, policies trained only on collected demonstrations struggle significantly under unseen lighting conditions. While Color Jitter provides modest improvement, it fails to capture photorealistic lighting variations at evaluation time. \method~(Ours) achieves the highest success rates across all three tasks, demonstrating that photorealistic lighting augmentation via video diffusion is more effective than standard 2D color augmentation.

\subsubsection{Background}
To evaluate background generalization, we test policy deployment across three distinct background scenarios. We use \textbf{RoboEngine}~\cite{yuan2025roboengineplugandplayrobotdata} as the augmentation-specific baseline, which segments objects and robot arms and applies 2D image inpainting to replace the background. As shown in Table~\ref{tab:augmentation_comparison}, \method~(Ours) achieves the highest success rates across all tasks. RoboEngine struggles in segmentation for certain objects, leading to frame-level inconsistencies and degraded gripper-object contact regions. Furthermore, its frame-by-frame processing cannot ensure temporal consistency or support multi-view generation, both of which \method naturally handles through video diffusion.

\subsubsection{Camera View}
To evaluate camera view generalization, we test policy deployment across four camera perspectives. We use \textbf{Fine-Tuned VISTA}~\cite{tian2024vista} as the augmentation-specific baseline, a diffusion-based novel view synthesis method that leverages ZeroNVS~\cite{sargent2024zeronvs} to augment third-person viewpoints from a single third-person view, which we extend to the bimanual manipulation setting. As shown in Table~\ref{tab:augmentation_comparison}, \method~(Ours) substantially outperforms VISTA across all three tasks, with particularly large margins on \texttt{Lift Roller} and \texttt{Place Cans in Plasticbox}, demonstrating the advantage of video diffusion-based multi-view generation over single-view synthesis approaches.

\subsubsection{Object Color}
To evaluate object color generalization, all baselines are trained on demonstrations with red objects and evaluated on gray objects at deployment time. %
We use \textbf{SAM3}~\cite{carion2025sam3segmentconcepts} as the augmentation-specific baseline, which segments objects of interest and applies color jitter to the segmented regions. A relevant prior work is ROSIE~\cite{yu2023scaling}, an object editing method, but its code is not publicly available, making SAM3 the most practical alternative. Similar to RoboEngine, SAM3 struggles to consistently segment small or distant objects, leading to temporal inconsistencies across frames and degraded quality at gripper-object contact regions. As shown in Table~\ref{tab:augmentation_comparison}, \method~(Ours) achieves substantially higher success rates, demonstrating the advantage of photorealistic video diffusion-based object color augmentation over segmentation-based color editing.

\subsubsection{Wrist + 3rd Person View}
To evaluate wrist and third-person view generation, we test policy deployment using a left wrist camera, right wrist camera, and a third-person camera simultaneously. Unlike the other augmentation dimensions, there are no suitable public baselines for this setting at the time of writing. Instead, we treat \textbf{ACT w/o Aug} trained on 100 real wrist-camera demonstrations collected directly with the target camera configuration as a performance reference, representing what is achievable with real collected wrist-camera data. As shown in Table~\ref{tab:augmentation_comparison}, \method~Pose-Only with 100 generated demonstrations already approaches the performance of ACT w/o Aug without collecting any real wrist-camera data. Scaling to 1000 generated demonstrations with \method~(Ours) further improves performance, achieving perfect success rates of 20/20 on \texttt{Lift Roller} and \texttt{Stack Bowls}, demonstrating the effectiveness and scalability of \method for multi-view camera generation.

\subsubsection{Cross-Embodiment}
To evaluate real-world cross-embodiment transfer, we collect demonstrations on a bimanual xArm7 and apply forward and inverse kinematics in MuJoCo simulation~\cite{mujoco} to retarget the trajectories to the bimanual Franka Panda. The retargeted demonstrations are then replayed in the RoboTwin simulator to generate Canny-edge control videos for the video diffusion model. We use a bimanual xArm7 as the source robot rather than the bimanual UR5 used in simulation, as we do not have access to WSG grippers in our real-world setup. These experiments complement our simulation results (Section~\ref{ssec:sim_results}) by validating cross-embodiment transfer on physical hardware with a different source robot. 
We compare against \textbf{Shadow}~\cite{lepert2024shadow} as our baseline, following the same bimanual adaptation described in Section~\ref{ssec:sim_results} for simulation experiments. Unlike Shadow, \method preserves photorealistic gripper and object contact regions, which is critical for precise manipulation tasks. As shown on the right side of Table~\ref{tab:cross-embodiment}, \method~(Ours) not only outperforms Shadow but also surpasses the collected demonstration baseline, demonstrating that our generated demonstrations can serve as a scalable and effective substitute for real target robot data collection.

\section{Conclusion}
We present \method, a scalable data generation pipeline for bimanual imitation learning that synthesizes photorealistic demonstrations across seven augmentation techniques via video diffusion conditioned on Canny-edge control videos, reference images, and language instructions. \method consistently outperforms augmentation-specific baselines in simulation and the real world, demonstrating that scalable data generation can substitute for costly real-world data collection. We hope \method inspires further work in video generation for robot learning. 

\section{Acknowledgments}
We thank Josh Shangguan, Kelly Wu, and Kyle Hatch for helpful discussions and writing feedback, the Physical Superintelligence Lab for lending equipment, and Weiduo Yuan and Tien Toan Nguyen for hardware support.

\bibliographystyle{IEEEtran}
\bibliography{main}

\clearpage
\appendix
\subsection{Simulation Task Details}
\begin{figure}[h!]
    \centering
    \includegraphics[width=0.48\textwidth]{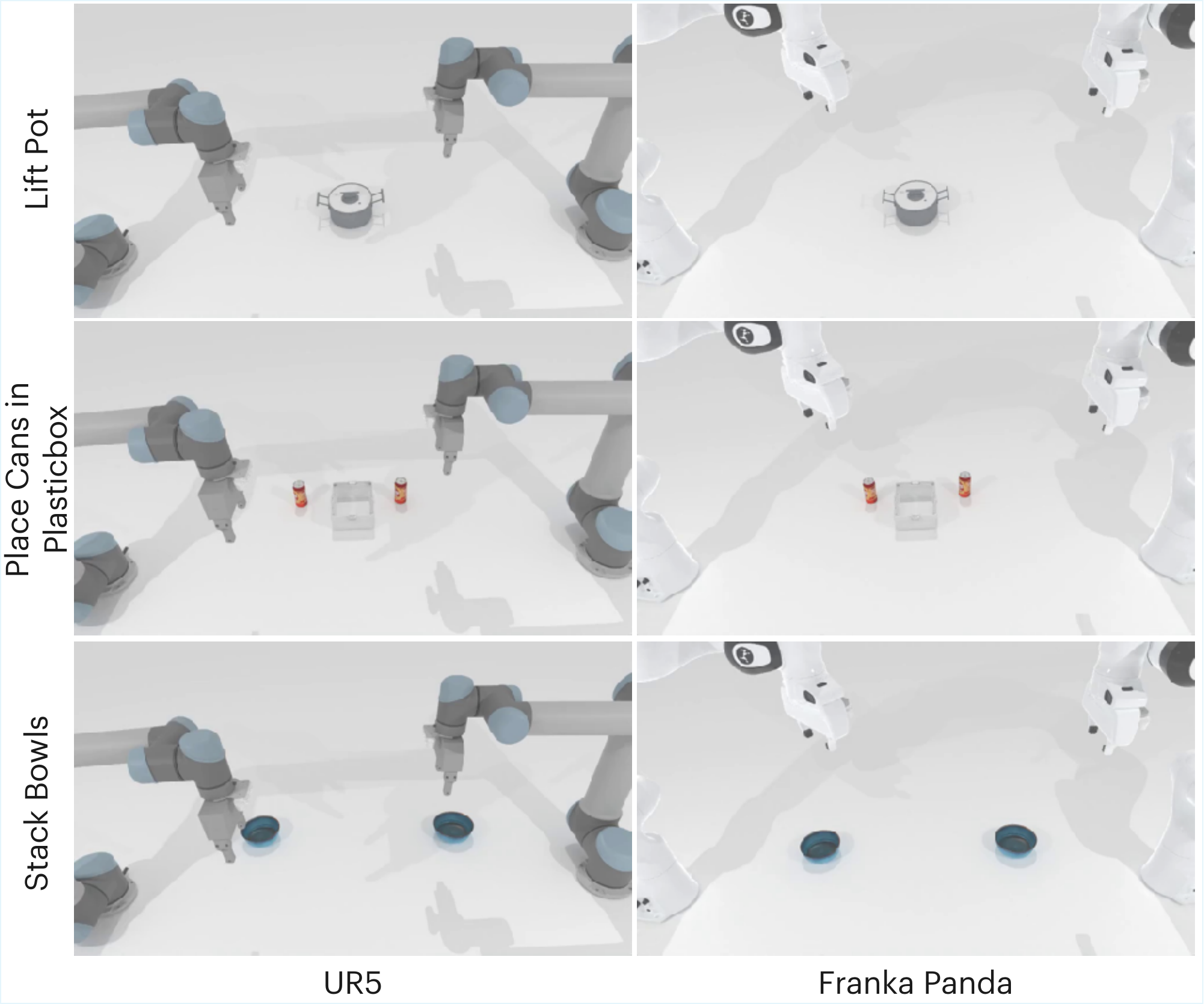}
    \caption{
        \textbf{Simulation Environments.} Bimanual manipulation tasks adapted from RoboTwin~\cite{chen2025robotwin}, shown for the bimanual UR5 with WSG grippers (left) and the bimanual Franka Panda (right). Tasks from top to bottom: \texttt{Lift Pot (LP)}, \texttt{Place Cans in Plasticbox (PC)}, and \texttt{Stack Bowls (SB)}.
    }
    \label{fig:simulation}
    \vspace{-5pt}
\end{figure}
For simulation experiments, we use RoboTwin as our simulator. We modify the original task setup, which uses both wrist and third-person cameras for policy training, to use only the third-person camera for experiments that do not require wrist-camera observations, improving ACT policy performance.

For cross-embodiment transfer, we use MuJoCo~\cite{mujoco} to perform FK and IK conversions from the xArm7 to the RoboTwin Franka Panda, as RoboTwin does not natively support the xArm7 at the time of writing. Specifically, we use \texttt{mash}, a MuJoCo-based retargeting tool developed internally in our lab, which is not publicly available at the time of writing. 

As DexMimicGen~\cite{jiang2025dexmimicen} did not release their trajectory expansion code at the time of writing, we implemented the trajectory expansion procedure as closely as possible to the method described in their paper.

\subsection{Task Selection Details}
We select tasks that span a diverse range of bimanual coordination strategies, demonstrating that \method generalizes across different task configurations. \texttt{Lift Roller} and \texttt{Lift Pot} are coordinated tasks requiring both arms to simultaneously grasp and lift an object with precise synchronization. \texttt{Place Cans in Plasticbox} is a parallel task where each arm operates independently to pick and place cans. \texttt{Stack Bowls} is a sequential task where subtasks must be completed in a specific order, requiring the policy to reason about task progression.

\subsection{Real-World Task Details}
The \texttt{Lift Roller} task uses a dough roller as the manipulation object. To initialize object poses in simulation, we use AprilTags~\cite{wang2016apriltag} to estimate the position of each object relative to the robot arms from the third-person camera, and set the corresponding object poses in the RoboTwin simulator accordingly. 

We exclude the simulation \texttt{Lift Pot} task from real-world evaluation due to safety concerns, as the task requires the grippers to operate in close proximity to the workspace table.
\begin{figure}[t]
    \centering
    \includegraphics[width=0.48\textwidth]{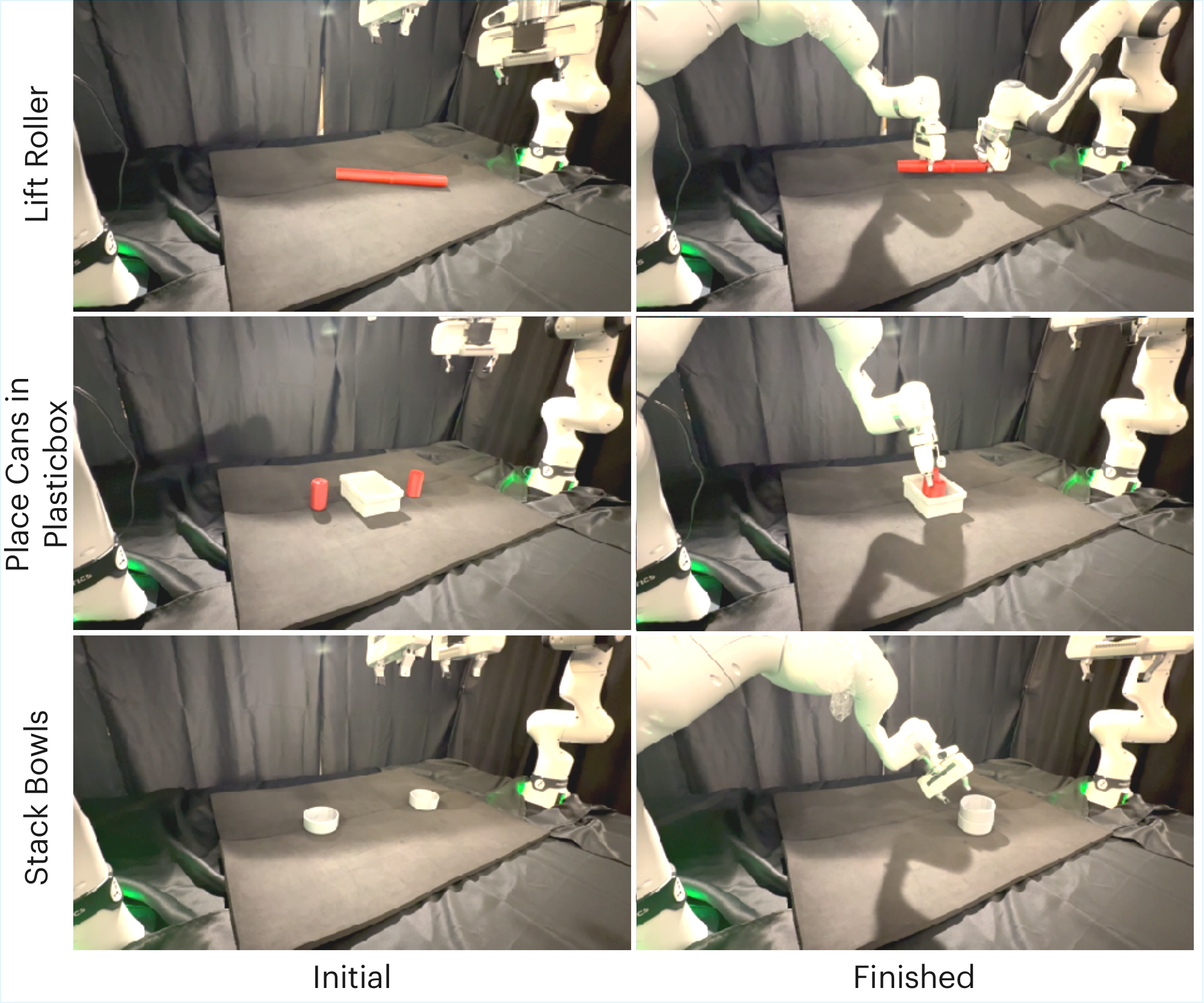}
    \caption{
        \textbf{Real-World Environments.} Bimanual manipulation tasks adapted from RoboTwin~\cite{chen2025robotwin}, shown for the bimanual Franka Panda. Tasks from top to bottom: \texttt{Lift Roller (LR)}, \texttt{Place Cans in Plasticbox (PC)}, and \texttt{Stack Bowls (SB)}. The left images show the initial state and the right images show the final states.
    }
    \label{fig:real_world_tasks}
    \vspace{-10pt}
\end{figure}

\subsection{Real-World Setup}
\begin{figure}[h]
    \vspace{-10pt}
    \centering
    \includegraphics[width=0.48\textwidth]{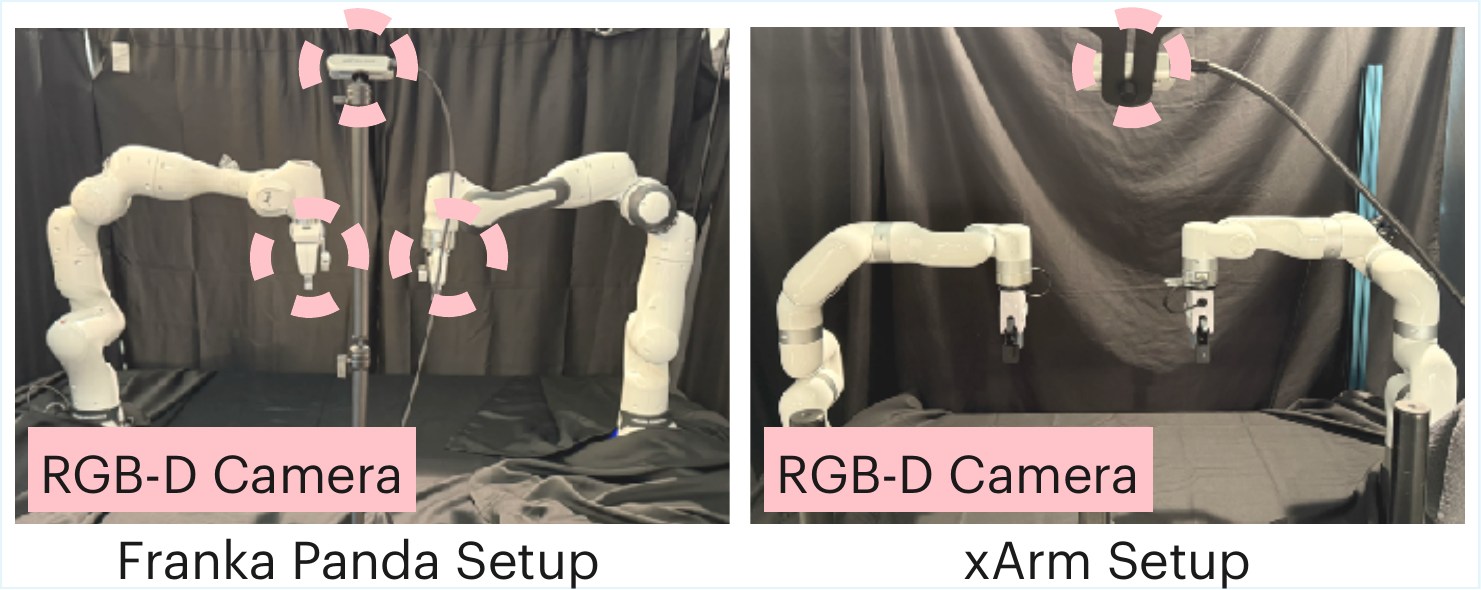}
    \caption{
        \textbf{Real-World Setup.} Bimanual cross-embodiment setup showing the Franka Panda (left) and xArm7 (right). Pink dotted circles indicate the Intel RealSense D435i camera placements.
    }
    \label{fig:real_world_setup}
    \vspace{-0pt}
\end{figure}
We leverage the GELLO~\cite{wu2024gello} setup to collect demonstration data for both the Franka Panda and xArm7 setups. 

\subsection{Reference Image Comparison}
\begin{figure}[h]
    \centering
    \includegraphics[width=0.48\textwidth]{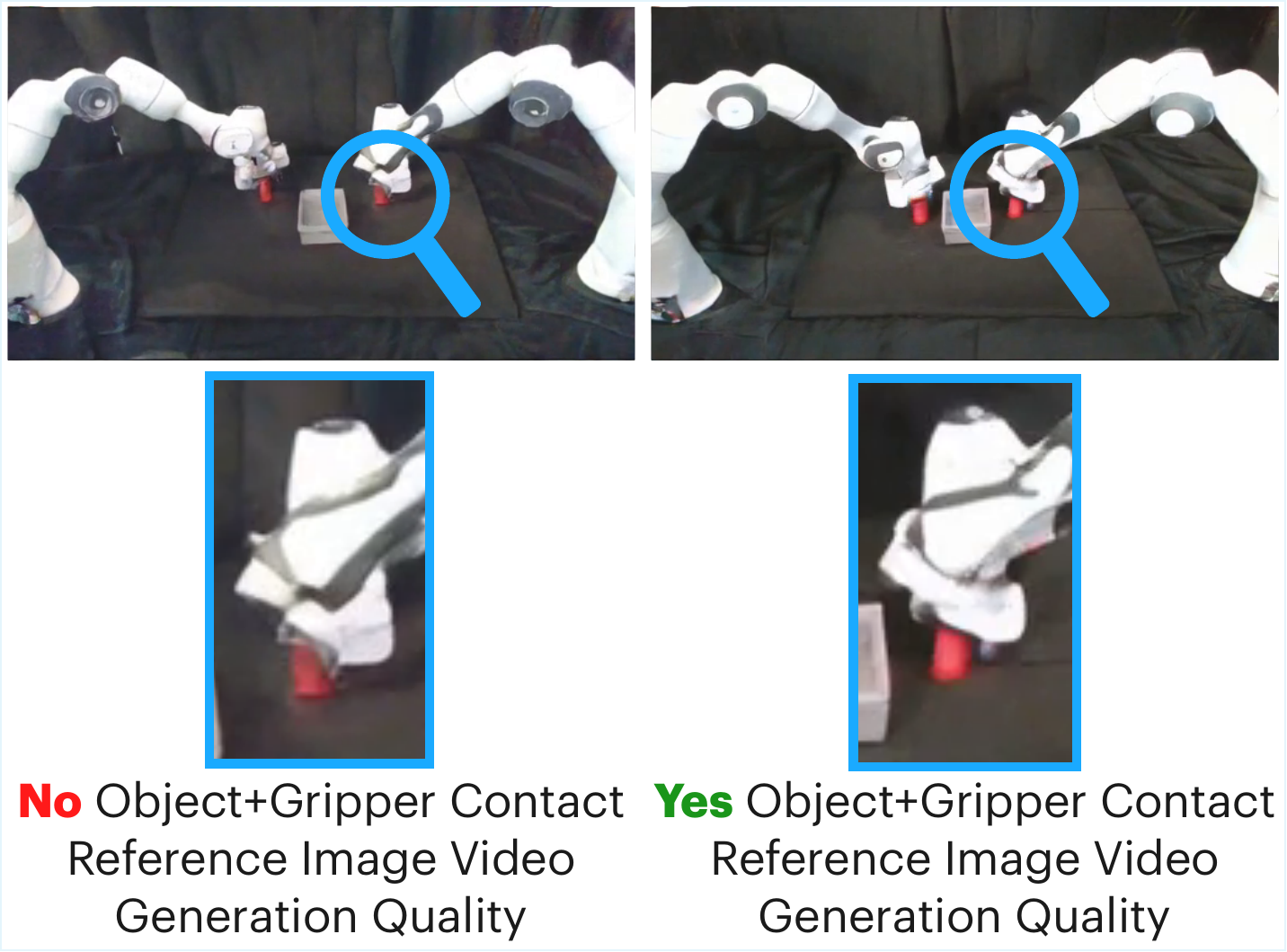}
    \caption{
        \textbf{Video Generation Quality Comparison.} Generated video frames without (left) and with (right) a reference image capturing gripper-object contact. As seen in the blue bordered images, including gripper-object contact in the reference image leads to higher fidelity contact synthesis in the generated video.
    }
    \label{fig:gripper_contact}
    \vspace{-10pt}
\end{figure}
We compare video generation quality when conditioning on a reference image that captures gripper-object contact versus one that does not. As shown in Figure~\ref{fig:gripper_contact}, including gripper-object contact in the reference image leads to noticeably higher fidelity contact synthesis and fewer visual artifacts on the robot arms. This is because Wan2.1-Fun-Control uses the reference image as a strong appearance prior, by providing an image that explicitly shows the gripper in contact with the object, the diffusion model can better infer the spatial relationship between the gripper and object, resulting in more realistic and geometrically consistent contact regions in the generated video.

\subsection{Additional Baseline Implementation Details}
Shadow~\cite{lepert2024shadow} is not publicly available so we implemented the baseline using the open-source code from their follow-up work Phantom~\cite{lepert2025phantomtrainingrobotsrobots} adapting it as closely as possible to the original Shadow method.

\subsection{Additional Real-World Experiment Details}
\subsubsection{Lighting}
\begin{figure}[h]
    \vspace{-10pt}
    \centering
    \includegraphics[width=0.48\textwidth]{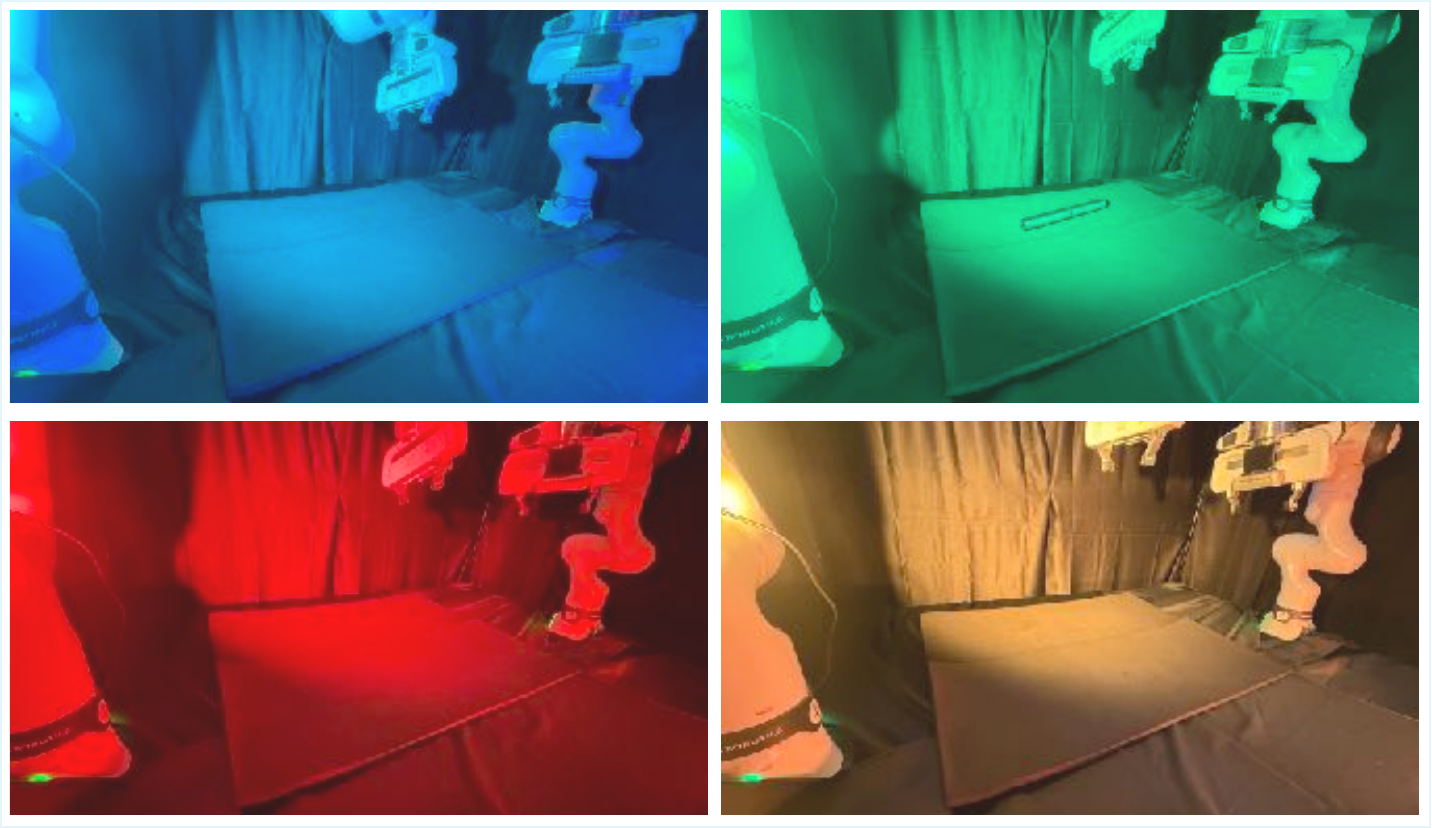}
    \caption{
        \textbf{Lighting Environments.} Test lighting conditions used for evaluation: blue (top left), green (top right), red (bottom left), and yellow (bottom right) ambient illumination.
    }
    \label{fig:lighting}
\end{figure}
For lighting, we leverage one NanLite Forza 60C and one NanLite PavoTube II 15C to display different colored lights in the scene. We show the different lighting conditions in Figure~\ref{fig:lighting}.

\subsubsection{Background}
\begin{figure}[h]
    \centering
    \includegraphics[width=0.48\textwidth]{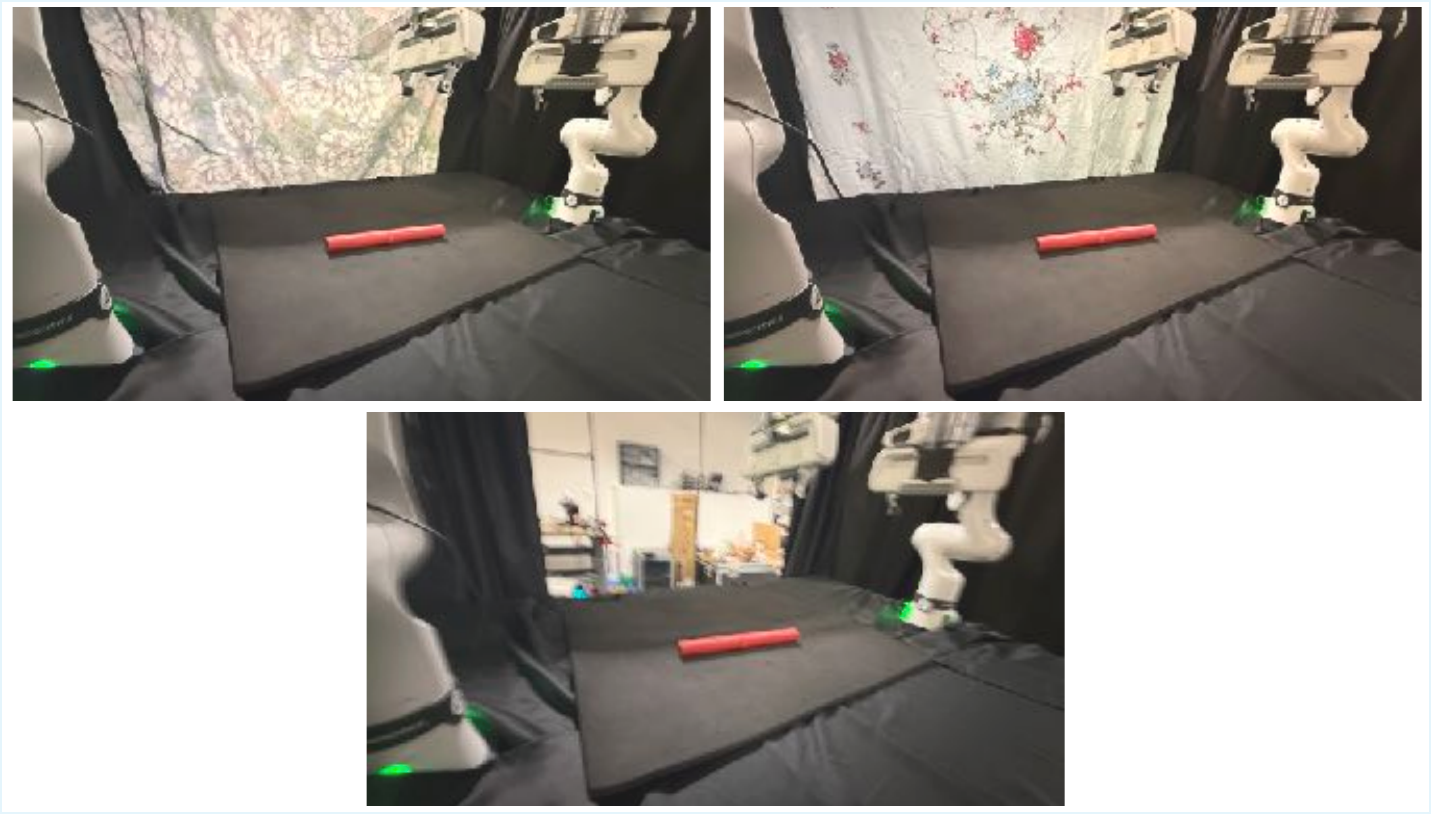}
    \caption{
        \textbf{Background Environments.} Test background conditions used for evaluation: flower fabric (top left), rose fabric (top right), and no curtain (bottom left).
    }
    \label{fig:background}
    \vspace{-10pt}
\end{figure}
For background generalization, we evaluate across three distinct background conditions: a flower fabric, a rose fabric, and an open lab environment with all curtains removed to simulate a cluttered real-world setting. We show the different background scenarios in Figure~\ref{fig:background}

\subsubsection{Object Color}
\begin{figure}[h]
    \centering
    \includegraphics[width=0.48\textwidth]{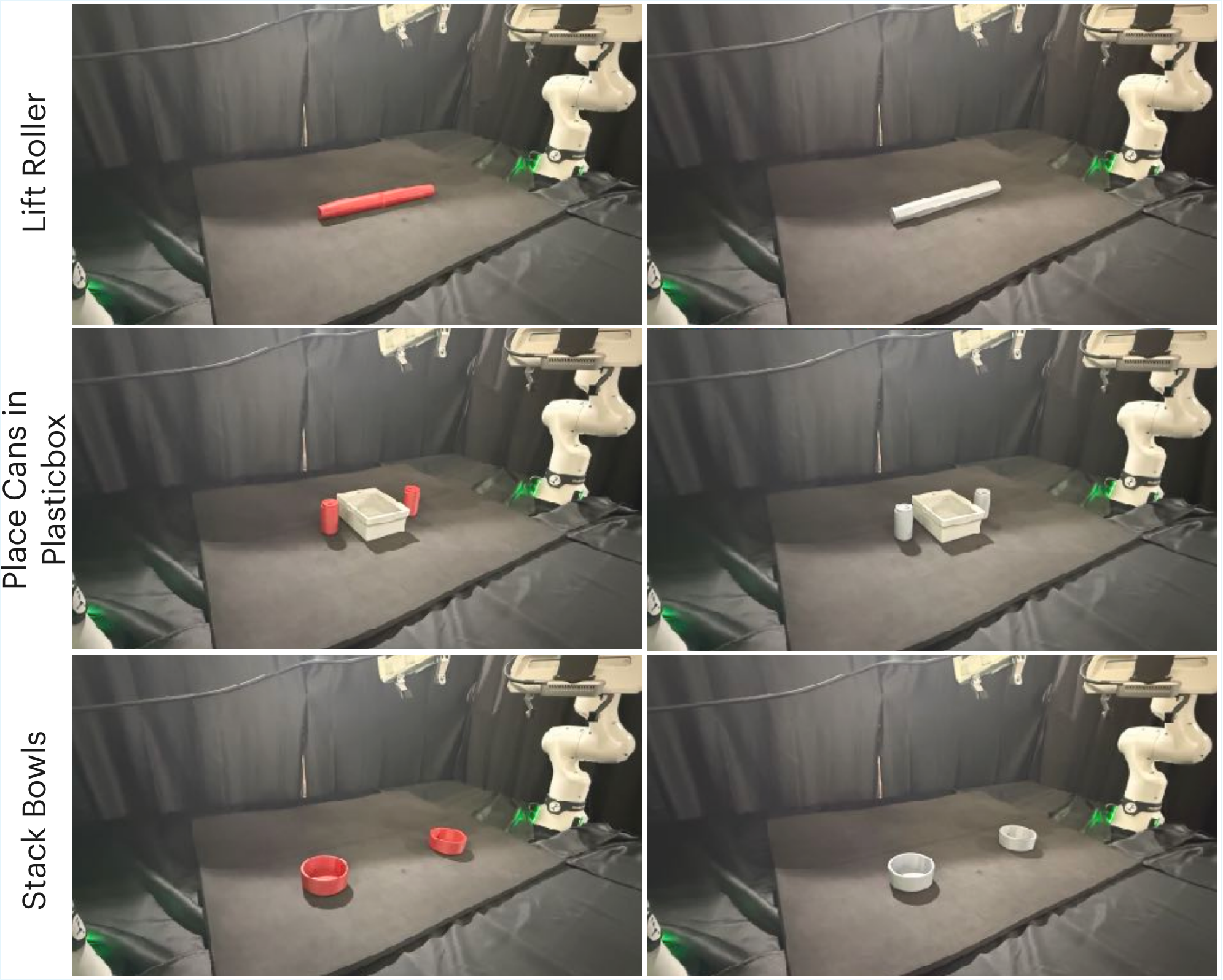}
    \caption{
        \textbf{Object Color Conditions.} Object color variations used for evaluation across each task. From top to bottom: \texttt{Lift Roller (LR)}, \texttt{Place Cans in Plasticbox (PC)}, and \texttt{Stack Bowls (SB)}.
    }
    \label{fig:object_color}
\end{figure}
For object color generalization, we evaluate across two object color variations. The objects were fabricated using a 3D printer, allowing us to vary color by swapping filament; we evaluate two colors to avoid unnecessary material waste. However, our method is not limited to only gray and red object colors. The two object color examples are shown in Figure~\ref{fig:object_color}.

\subsection{Canny-Edge Filtering}
\begin{figure}[h]
    \centering
    \includegraphics[width=0.48\textwidth]{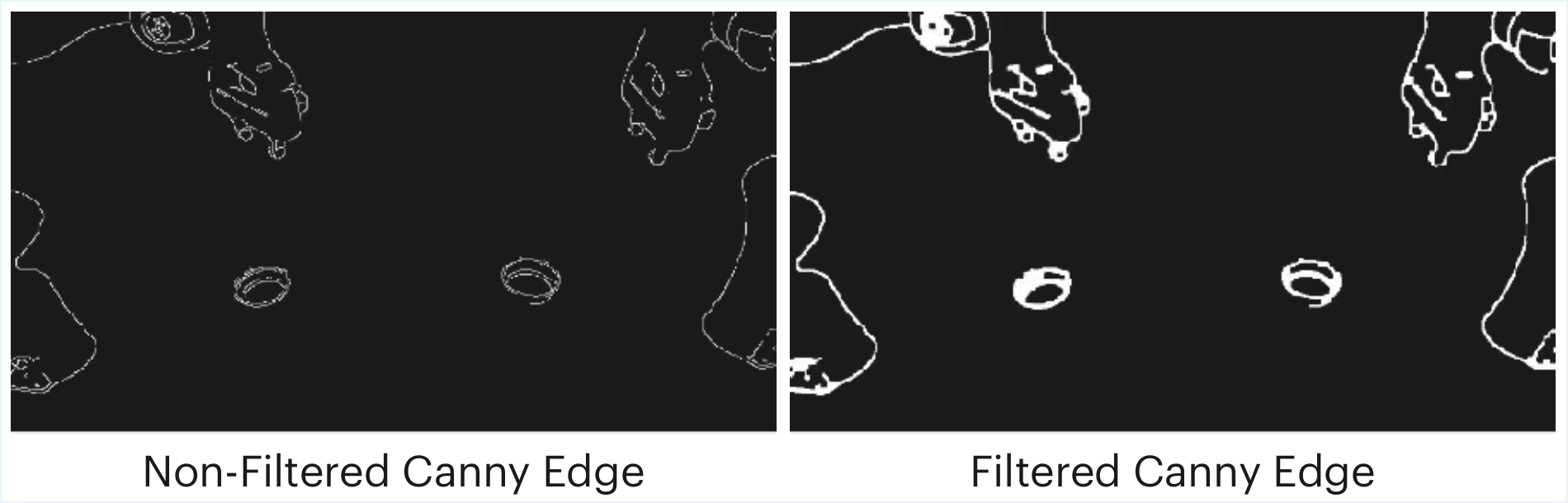}
    \caption{
        \textbf{Canny-Edge Filtering Comparison.} Unfiltered (left) vs. filtered (right) Canny-edge representations, showing how edge thickening and connectivity post-processing produce cleaner structural control signals for the video diffusion model.
    }
    \label{fig:canny-filtering}
    \vspace{-10pt}
\end{figure}
Rather than using raw Canny-edge outputs, we apply two post-processing steps to produce cleaner and more informative control signals: (1) edge thickening to strengthen the structural control signal, and (2) edge connectivity to bridge nearby disconnected edges and reduce fragmented edge artifacts. These steps ensure that the Canny-edge control video provides clear and coherent structural guidance to the video diffusion model. We show an example of the non-filtered and filtered canny-edge comparison in Figure~\ref{fig:canny-filtering}.

\subsection{Limitations and Future Work Opportunities}
\begin{enumerate}
    \item Like all video generative model-based approaches, synthesized videos may contain visual artifacts or temporal inconsistencies that hinder downstream policy learning.
    \item The third-person camera must be positioned close to the robot and objects of interest, as distant views produce noisy Canny-edge representations that degrade generation quality, particularly at gripper-object contact regions.
    \item Although video generation is performed zero-shot, achieving high-quality results requires careful prompt engineering and informative reference images.
    \item \method's trajectory expansion procedure requires access to a simulator and object meshes to construct a digital cousin, similar to DexMimicGen~\cite{jiang2025dexmimicen}. While this is a shared assumption, it may limit applicability to tasks or objects that are difficult to simulate accurately.
    \item \method assumes tasks can be decomposed into object-centric subtasks for trajectory expansion.
    \item \method has not been tested on deformable objects however future work could leverage recent approaches such as SoftMimicGen~\cite{moghani2026softmimicgendatagenerationscalable} to extend it in this direction.
\end{enumerate}

\end{document}